\documentclass{article}
 
\usepackage{graphicx}
\usepackage{fullpage}
\usepackage{url}

\begin{document}

\title{Image and Information\footnote{This column is a translation of a french article I wrote for the french scientific association {\em La main \`a la p\^ate} (\protect\url{http://www.fondation-lamap.org/}).
Following the book projects ``Matter and Material''~\cite{MM-2013} (2013) and ``One energy, any energy''~\cite{EEs-2015} (2015) published by Belin, the french column ``Image et Information'' is
 intended to be published in the book on {\em ``Information and Communication''} (2017). 
The targeted audience are teaching professors or assistants in secondary schools (K-12 in the US, and potentially in primary schools) that teach children aged from 6 years old to 15 years old. 
}
}

\author{Frank Nielsen\\ \'Ecole Polytechnique\\ Sony Computer Science Laboratories, Inc.\\ \ \\ E-mail: {\tt Frank.Nielsen@acm.org}}
\date{February 2016}

\maketitle

\begin{abstract}
A well-known old adage says that {\em ``A picture is worth a thousand words!''}
(attributed to the Chinese philosopher Confucius ca 500 years BC).
But more precisely, what do we mean by information in images? And how can it be retrieved effectively by machines?
We briefly highlight these puzzling questions in this column.
But first of all, let us start by defining more precisely what is meant by an ``Image.''
\end{abstract}

\section{The broad notion of Images: From digital photos to spatio-temporal reconstructed images}

\subsection{Images, pictures and highly compressible photos}

A first quick preliminary answer to define an image  in the digital era is to say that an image is merely a {\em picture}.
A digital picture is represented by a {\em grid of pixels} $I(x, y)$, where at each position $(x, y)$ of the grid, a picture element ({\em PIXEL} = PIcture X ELement) stores a color RGB triplet: $I(x, y)=(\mathrm{red}, \mathrm{green}, \mathrm{blue})$. 
The dimension of the image is the number of lines $L$ times its number of columns $C$ times the number of color channels. 
Each {\em color channel} of a pixel stores an integer value, usually between $0$ and $255$ (coded in binary by $8$ bits), but there exists also high-resolution color images with 
$14$-bit coded values or even more. 
Indeed, the light sensors in the cameras convert the aggregated amount of light received during a time lapse (the exposure time) in an electrical response by photodiodes. These electrical signals are then discretized to form a digital photo. 
For uncompressed photograph images, the size of such a raw image is very large because it requires $L \times C \times 3 \times 8$ bits to store and manipulate. 
The most popular formats for image compression are the TIFF and PNG standards. 
These formats get good compression ratio for {\em synthetic pictures} (created by a drawing program) but are not very effective for natural images: the {\em photo pictures}! 
Photo images of a natural scene  captured by sensors under slight noise condition are almost indistinguishable perceptually by the human eye. 
It is therefore better to store images by compressing (with a small but controlled loss): This is precisely the objective of the JPEG standard.

\subsection{Images are regularly distributed spatio-temporal data obtained by raw signal reconstruction}

\begin{figure}%
\centering
(a) \includegraphics[width=0.35\columnwidth]{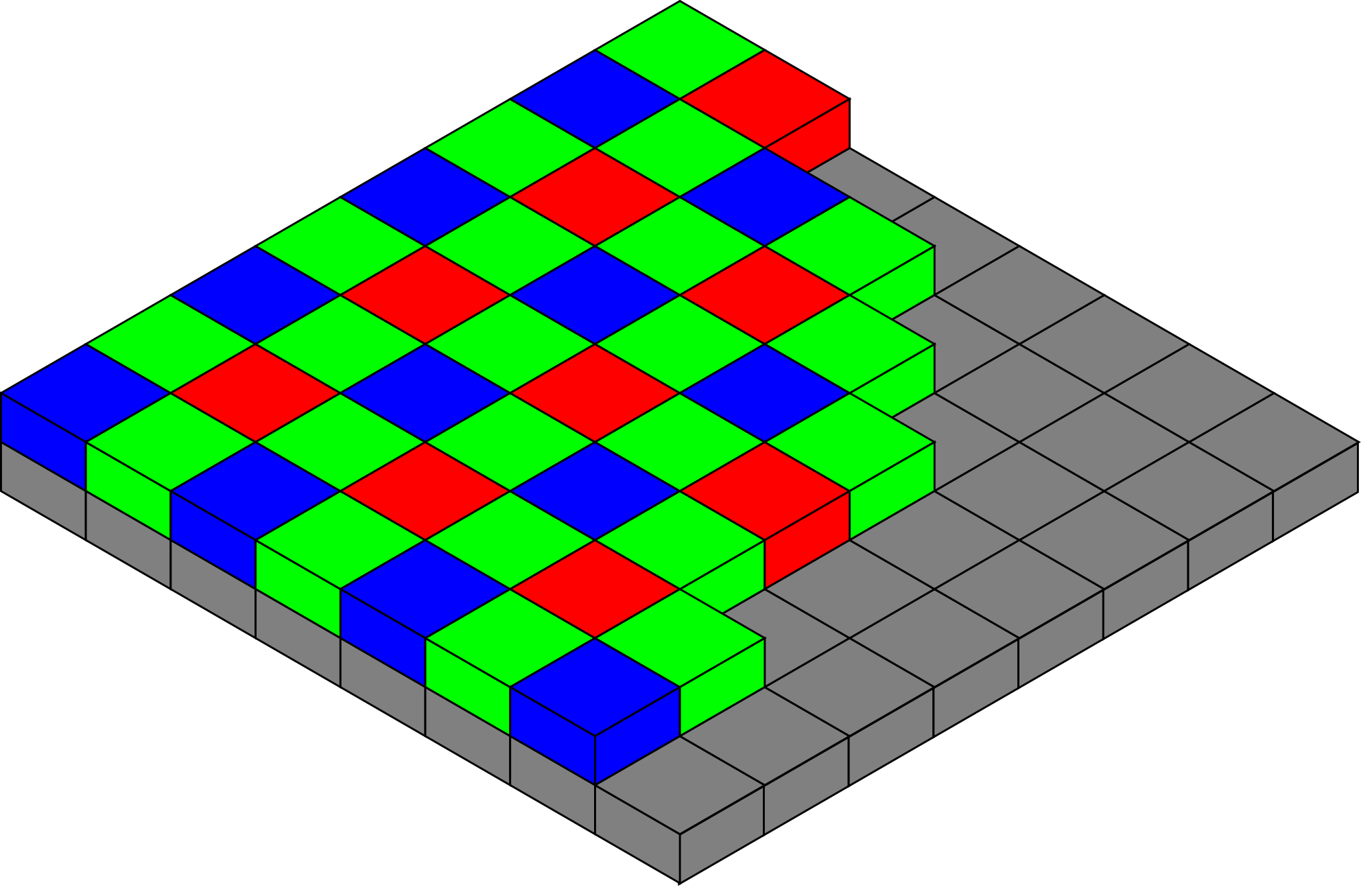}

\begin{tabular}{|l||c|c|c|}\hline
\includegraphics[width=0.2\columnwidth]{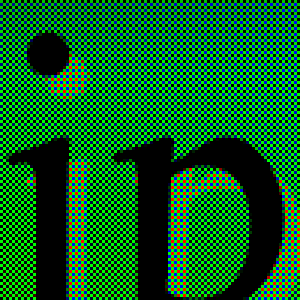} &
\includegraphics[width=0.2\columnwidth]{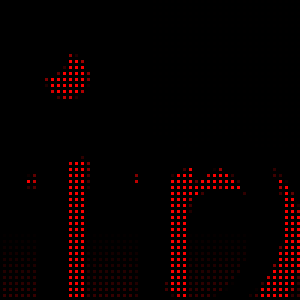} &
\includegraphics[width=0.2\columnwidth]{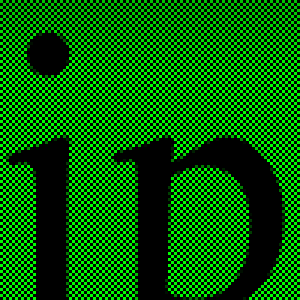} &
\includegraphics[width=0.2\columnwidth]{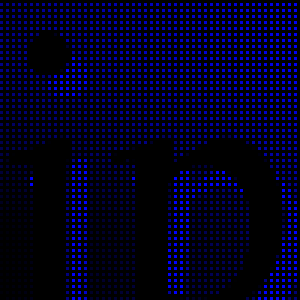} \\ \hline
(b) captured image & (c) red channel & (d) green channel & (e) blue channel\\ \hline
\end{tabular}

\vskip 0.2cm

\begin{tabular}{|c|c|}\hline
\includegraphics[width=0.15\columnwidth]{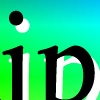} &
\includegraphics[width=0.15\columnwidth]{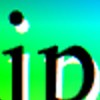} \\ \hline
(f) original image & (g) reconstructed image \\ \hline
\end{tabular}

\caption{An image (a) is a grid of (red, green, blue) color (RGB) pixels. 
A color image (f) can be  captured with a single light sensor overlaid with a Bayer filter ((a), a color filter array), and reconstructed from the red, green and blue channels (c), (d), (e) with an appropriate demosaicing algorithm (g).
Notice the loss of resolution and the artifacts located at edges (color fringes).
Images are courtesy of Wikipedia.
\copyright{} Wikipedia.
\label{fig:bayer}
}
\end{figure}

To reduce costs or to miniaturize digital consumer cameras, industrial companies do not use three sensors (one for each RGB channel) but rather a single sensor which is coated with a Bayer filter. The color image is then reconstructed through a process called demosaicing (see Figure~\ref{fig:bayer}). 
So even your selfies taken with mobile phones are actually {\em reconstructed images}! 
An image can be interpreted as a discrete signal: A discrete function sampled at regular positions of the pixels of the continuous density function which measures the amount of light received at a point on the sensor surface.

\begin{figure}
\centering

\begin{tabular}{|c|c|}\hline
\includegraphics[width=0.45\columnwidth]{ultrasoundimage.jpg} 
&
\includegraphics[width=0.45\columnwidth]{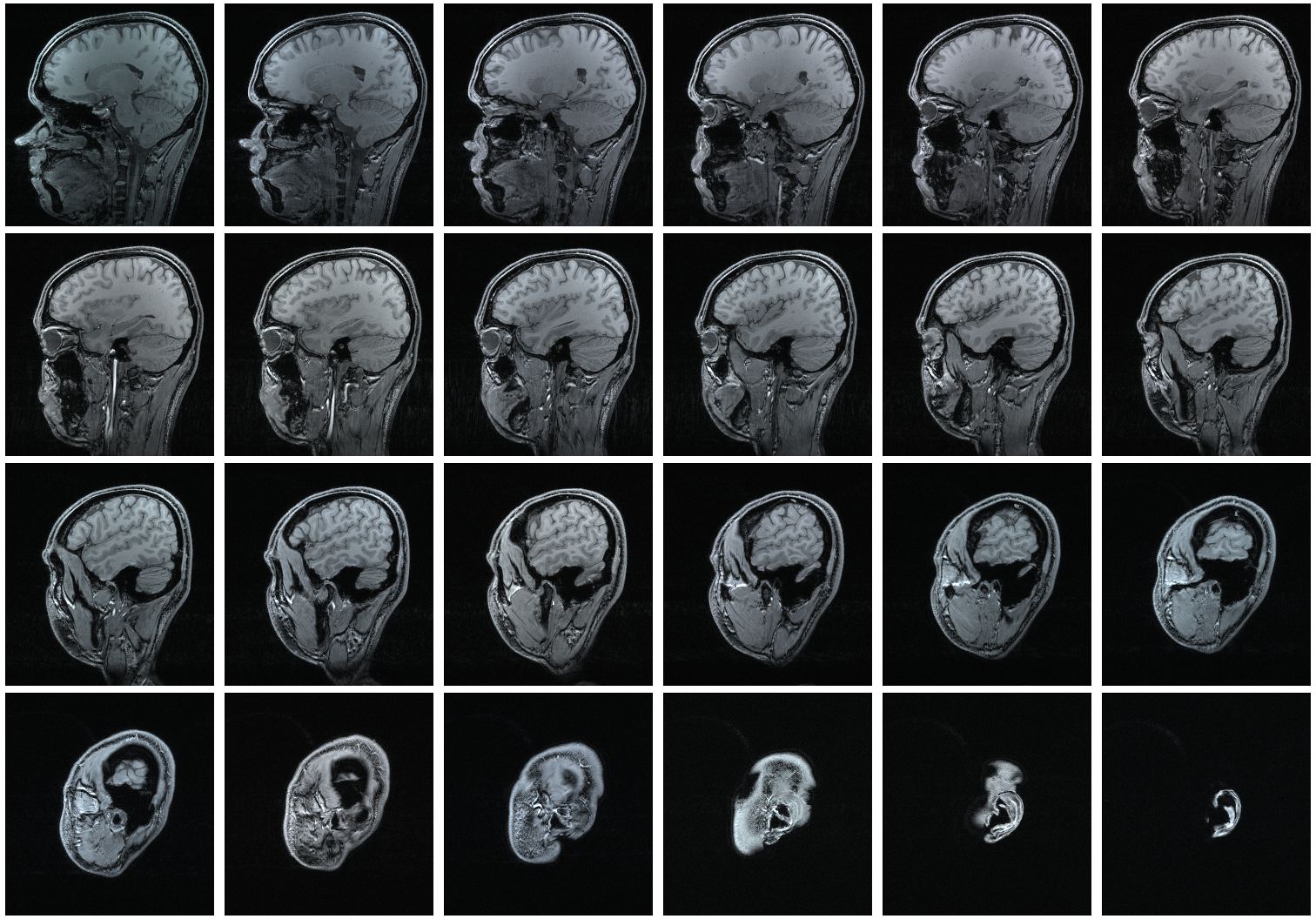}
\\
(a) ultrasound image & (b) MRI image\\ \hline
\end{tabular}

\caption{\label{fig:ultrasound} An ultrasound image (a) and a (b) MRI image (magnetic resonance imaging series of a human head)  obtained by reconstruction techniques from discrete raw signal data acquired by imaging sensors.
\copyright{} Wikipedia.}

\end{figure}

Although digital pictures (often shared in social networks) constitute a large proportion of image data ($140$ billion images on Facebook with $300$ million new images  uploaded every day), the very notion of ``Image'' is much larger than that of a digital photo picture as we will see from  now on! 
For example, in medical imaging, we encounter many types of images such as ultrasound images reconstructed from discrete signals obtained by tailored sensors, 
and not anymore directly optically observed as in the case of photos. 
These ultrasound images are highly useful, e.g., for diagnosing the development of the fetus in pregnant women (Figure~\ref{fig:ultrasound} (a)).
An {\em imaging modality} is a term used to refer to the acquisition technique used by sensors and the reconstruction method employed to output a reconstructed image from raw sensor data.

An image is not necessarily either a 2D grid of pixels, it can also be {\em multi-dimensional}: 
For example, Magnetic Resonance Imaging (MRI) produces 3D grid of voxels (voxel stands for VOlume X ELement) to visualize the various organs {\it in vivo} ((Figure~\ref{fig:ultrasound} (b)). 
For these reconstructed data, one uses the term of {\em pseudo-colors} instead of colors, 
and image technicians can interactively define a pseudo-color palette to highlight such or such part of the image  by manipulating a transfer function that maps raw data to pseudo-colors for displaying on a monitor. 
An image is not necessarily  a square either! 
For example, radarscope images in airport control towers that monitor aircrafts have disk-shaped visualization {\em support}
(and reconstructed using the physical principle of the Doppler effect).

An image can also have more than three standard RGB channels: 
For example, in {\em hyperspectral imaging}, one can deal with several hundreds of channels! 
These hyperspectral images are often obtained by satellites that monitor the Earth, and help classify land by distinguishing urban from rural areas, and in forests, allow one to obtain the geographical spread of the different tree species, etc. 
The {\em spectral image} divides the {\em electromagnetic spectrum} in {\em bands} of wavelengths (corresponding to a channel per band) which may also be located outside the visible range of the human eye (say, in the infrared or ultraviolet ranges) .
Finally, images may not be {\em static} but can also be {\em dynamic}, that is, changing with time.
A {\em video} can be interpreted as an abstract 3D image, with the third dimension denoting the discrete time. 

\begin{figure}[b]
\centering

\includegraphics[width=0.8\columnwidth]{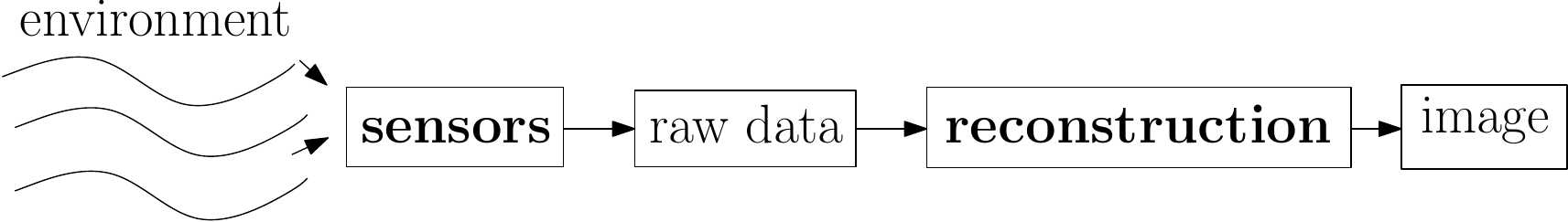}

\caption{Pipeline of an image modality: Sensors capture the environment and discretize the signal in raw data that is then processed by a reconstruction algorithm to output an (abstract) image. \label{fig:abstractimage} 
}
\end{figure}

To conclude with a modern definition of an Image, we may say that an image is a space-time discrete signal regularly sampled.
Figure~\ref{fig:abstractimage} depicts a conventional pipeline for acquiring an image.

It is interesting to build new image modalities:
For example, this is an active research topic  to capture the activity of living cells in biology to better understand underlying   scientific phenomena.
Professor Eric Betzig received the Nobel prize\footnote{\url{http://www.nobelprize.org/nobel_prizes/chemistry/laureates/2014/betzig-facts.html}} in 2014 ``for the development of super-resolved fluorescence microscopy''.

\subsection{The ultimate photographic image:  Capturing and visualizing the 7D plenoptic function}

We have explained that a photo image is a discrete function regularly sampled at positions of a 2D grid that stores information on the spectrum of the amount of light received during an exposure time, and converted into the RGB channels for later processing convenience. 
For a long time, digital photography\footnote{The first digital camera was developed in 1975 by Steven Sasson, an engineer at Kodak.} had to catch up the imaging capabilities of an analog silver picture.\footnote{First silver photo obtained by french Nic\'ephore Ni\'epce in 1827.}
Nowadays, digital photos have significantly exceeded capabilities of conventional film photos: 
Digital image have ultra high-resolution  (close to terapixels), full field of view ($360$-degree panoramic images), etc.
Defining the very notion of a digital photo is an active research topic, 
and many novel avenues are currently being explored in order to redefine the  experience to capture a scene or a moment.

The {\em plenoptic function}~\cite{PlenopticFunction-1991} (``plenus'' in latin means complete) defined for each point $(X, Y, Z)$ of space, for each orientation of the light ray $(\theta, \phi)$ anchored at this point, for each given wavelength $\lambda$ of the electromagnetic spectrum and for each given time $t$, a real value. 
The plenoptic function $f(X,Y,Z,\theta,\phi,\lambda,t)$ is seven-dimensional (7D), and allows one thus to capture the complete light field in the vacuum space. 
$360$-degree panoramic images (surround images) can be interpreted as sub-functions of this plenoptic function, with each panorama being determined the given fixed position $(X, Y, Z)$ of the optical center of the camera. 
Recent companies (like Lytro, Raytrix, etc.) have entered the digital consumer photo market by proposing lightfield cameras that capture  4D light fields conceptually as follows: 
Think of a traditionnal light sensor, and replace each pixel by a small 2D image sensor.
These 4D lightfields allow one to enjoy  novel applications,  like   reworking the focus of an image after it has been captured.
We will now focus on the notion of information contained in an image, and the computational processes to distill it.
This is covered academically by the fields of {\em image processing} and of {\em computer vision}.

\section{What is the information concealed in an image and how to extract it?}

\subsection{Visual information processing:  Brain perception versus artificial intelligence} 

The human brain is the intelligent organ that integrates the human body information, and generates the perception of our five senses (sight, hearing, smell, taste and touch) of our surrounding environment. 
With  $100$ billion interconnected neurons, the human brain is capable of unequaled flexible intelligence.
 Much of this neuronal activity is devoted to processing the visual information that starts from light sensor cells located in the retina of the eye: cones (three types of cone play the role of the three   RGB channels) located in the center of the retina, and the rods located at the periphery that  respond more quickly and thereby are more effective to detect motion. 
Unlike planar sensor of video cameras, note that the eye sensor cells do not form a regular grid, are  positioned spherically on the retina, and time is not regularly discretized! 
It has been shown that human perception generated by the visual processing and consciousness are the result of the neuron activity:
NCC for Neuron Correlates Consciousness.

 It has been shown experimentally that at the low level of the visual processing chain, the neuron activity is performing {\em filter operations} by separating the  incoming visual flow into frequency bands: 
The {high frequencies} components of images correspond to edges or contours of image objects  and those {\em high-frequency information} are very well perceived closely (say, like the moustaches of cats). 
As we increase the viewing distance, one perceives rather the {\em low-frequency information}, the image texture. 
We can check and also use this perceptual phenomenon  with  an experience that consists in swapping the high frequencies of an image with the low frequencies of another image: We obtain so-called {\em hybrid images}~\cite{HI-2006} with two levels of perception:  One perceived image at near viewer position and
another perceived image at far viewer position (see  \url{http://cvcl.mit.edu/hybridimage/}).

\begin{figure}%
\centering
\includegraphics[width=0.35\columnwidth]{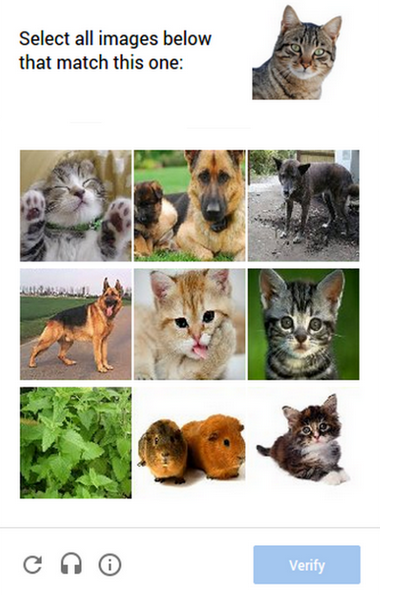}%

\caption{A visual CAPTCHA, to distinguish between high-level AI Humans from low-level AI Computers,   asks users to classify images according to a given image.\copyright{} Wikipedia.  \label{fig:captcha}}
\end{figure}

While making tremendous progress, machines have not yet the visual understanding capabilities of Human, and this rather a reassuring statement that is being used to verify that a user connected remotely to a machine is actually a human being (and not a machine that imitates a human) by CAPTCHAs~\cite{CAPTCHA-2003} that 
stands for {\em ``Completely Automated Public Turing test to tell Computers and Humans Apart.''}
For example, when creating a new user login for accessing web services, there is a visual CAPTCHA challenge to read that we need to pass in order to proceed (often,
visual CAPTCHAs ask to decode the text information encoded in deformed images).
There are various CAPTCHAs' systems nowadays.
For example, the CAPTCHA of Figure~\ref{fig:captcha} asks to select all images related to a given image.
This is an image classification task in computer vision.

\subsection{Photogrammetry: 3D reconstruction and geometric measurements}

\begin{figure}%
\centering
\includegraphics[width=0.45\columnwidth]{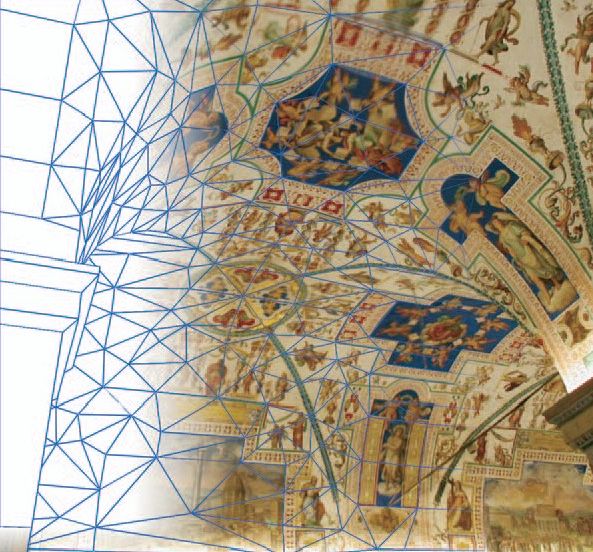}%

\caption{3D reconstruction of the indoor part of Vatican's celebrated library: From a collection of photo pictures, calibration is performed to recover the camera intrinsic parameters and the 3D location and gaze of the camera positions, the extrinsic parameters. Image features are then extracted and feature matching is performed to build a 3D mesh that can then be textured.
The left portion displays the raw 3D mesh, the middle part the textured mesh, and the right portion the captured image.
\copyright{} Frank Nielsen \& St\'ephane Nullans (cover image of~\protect\cite{b-n-visualcomputing-2005}).
}%
\label{fig:reconstruction}%
\end{figure}
With the spread of large-resolution digital photos and the computing power of machines, fully automatic 3D reconstruction of buildings and objects was developed successfully  during the last three decades.
The theory of multi-view reconstruction is built on the framework of projective geometry and constraints of positioning ideal pinhole cameras (sensors) for acquiring the different pictures. 

It is also possible from a pair of stereo images to use the {\em parallax information} to calculate the {\em geometric depth information} in a RGBZ picture, a 
{\em depth picture} or {\em disparity map}: 
Each pixel stores the color triplet RGB    and a depth information, $Z$ value. 
This kind of image  is termed {\em 2.5D image} because for each pixel, knowing its corresponding light ray, one can recover the 3D point from the $Z$ value~\cite{b-n-visualcomputing-2005}. 
However, one can not get a complete 3D reconstruction of the scene with a single RGBZ image because of potential occlusion not imaged.
Indeed, it is only the depth information of the first point encountered along a light ray that can be recovered. 
This field of {\em photogrammetry} also allows one to reconstruct high-precision image textures (say, building facades) by combining and integrating the various  corresponding portions in the captured images. 
With the rise of drones, 3D capture becomes inexpensive, automatic and robust.
The smartphone is also becoming a handy scanning device, say for capturing one's face and creating his/her avatar for realistic game play.

\subsection{Computer vision: From industrial robotics to smart cameras}

\begin{figure}%
\centering
\includegraphics[width=0.6\columnwidth]{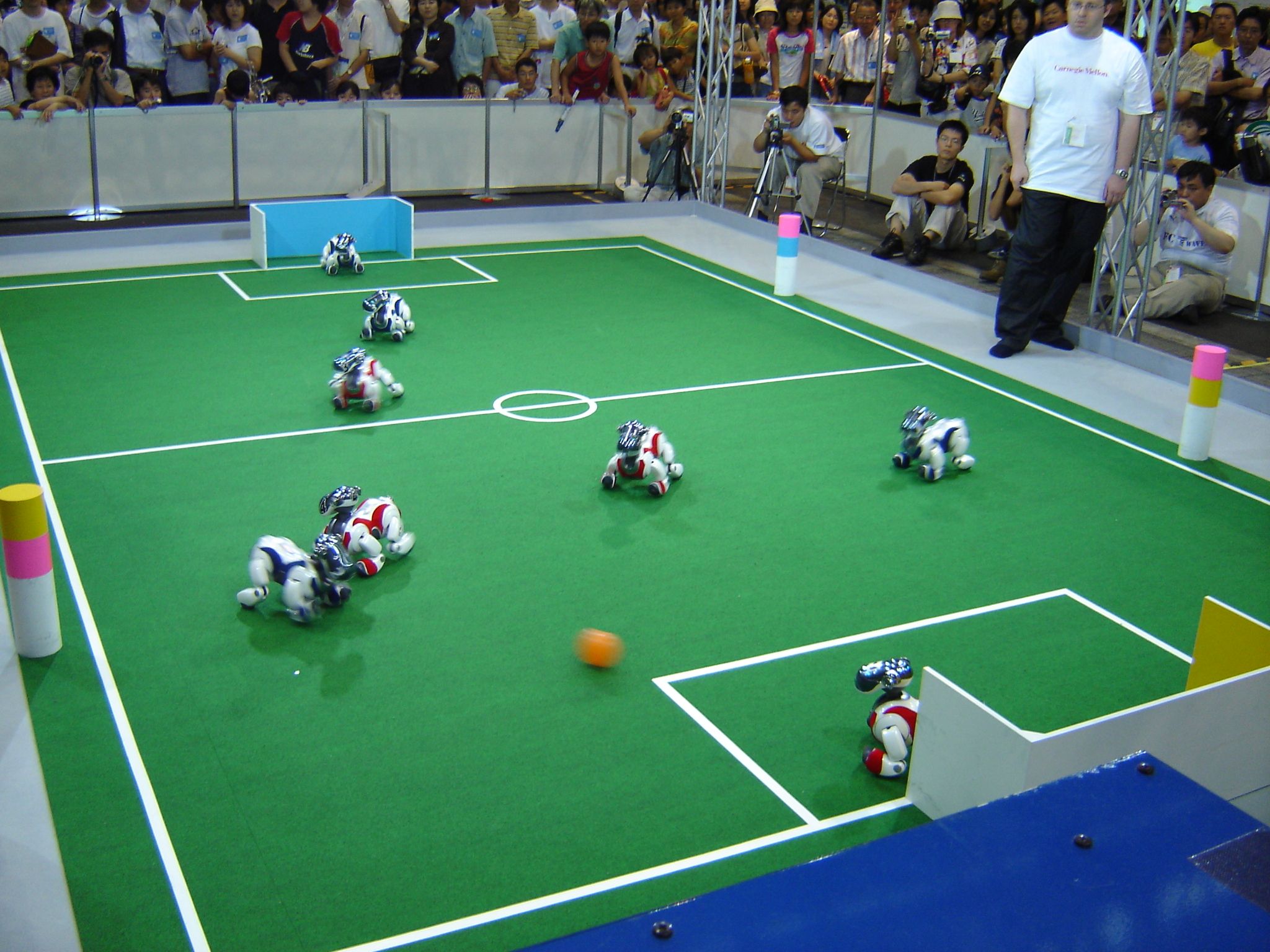}%

\caption{Robots Aibos playing soccer at Robocup in 2005.
\copyright{} Wikipedia. 
}%
\label{fig:robotcup}%
\end{figure}

\begin{figure}%
\centering
\includegraphics[width=0.6\columnwidth]{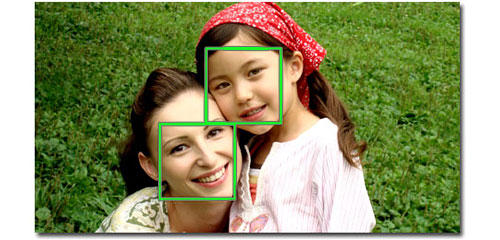}%

\caption{Face detection in photos and automatic shutter mode when everyone smiled.
\copyright{} Sony's ``smile shutter.''}
\label{fig:smileshutter}%
\end{figure}

Automation by articulated arm robots has fostered the development of {\em machine vision} techniques to  check the quality of manufactured products. Robotics and remote surveillance cameras have also pushed the envelope of intelligent video algorithms. 
For example, RoboCup\footnote{\url{http://www.robocup.org/}} is an annual international competition of robot teams playing collaboratively soccer. 
A basic task for these robots equipped with cameras is to locate the ball in the image (a red ball has been chosen for Robocup to ease the recognition task, see Figure~\ref{fig:robotcup}). 
This visual task is solved by {\em image segmentation} techniques where the image is partitioned into groups of pixels (literally the ``segments'') representing semantic objects in the image. That is, the segmentation returns the pixels of segments corresponding to the ball, the pixels corresponding to the ground, and the pixels of the  robot opponent, etc.). 
In video surveillance, the aim is to detect abnormal behavior or to characterize the activities of persons. 
Digital cameras have standard algorithms becoming increasingly sophisticated by combining {\em vision techniques} for feature extraction in the image 
with {\em learning techniques} on these characteristic data to recognize situations.
Those feature extraction plus machine learning algorithms can reliably recognize human faces and emotions in the photographs (and trigger the shot when everyone smiled for example, see Figure~\ref{fig:smileshutter}) or the type of scene (city skylines, fireworks, fog weather, etc.) and apply the corresponding post-processing after acquisition to obtain the best image (for example, by attenuating the fog or the rain to enhance the scene visibility). 
These visual algorithms are numerous and require a true expertise to extract from raw data (millions of image pixels) a smaller number of  features (say, a few thousands) which can then be used by more sophisticated algorithms for more complex visual learning tasks, such as automatically annotating the objects on a photo (with potential occlusion), or automatically generating text summaries for a given image (for example,  ``car stopped at a red traffic light, a pedestrian crossing with a stroller''). Understanding the images becomes a major  industrial challenge with the arrival of automatic driving cars.
To make these  control systems more robust, many sensors equip cars (like lasers complementing stereo camera rigs) and the task is to design algorithms that {\em integrate} all these information for
 automatically making decision (for example, to know when to slow down or speed up). This field bears academically the name of {\em information fusion}. 
The availability of large data sets (big data) allows statistical  algorithms to perform well by examining many situations encoded inside the data automatically by learning machines (and not programming them case by case by human programmers).
Nevertheless, those algorithms are still not perfect: Processing visual information   can not generally be $100\%$ certified, and those algorithms must also measure the risk, and for example, enable the driver to take back the control of the car in novel unseen (risky) situations.

\subsection{The new perspectives offered by deep learning: Explicit versus implicit features}

Traditional computer vision algorithms described above is based on knowing what kind of features to extract from the image. 
This is a very difficult and tricky task performed manually by engineers. 
So instead that or that features from a catalogue, programmers tend to extract many different types of features, 
and then try to find a way to {\em select} and  {\em combine} effectively those features. 
Therefore, traditional visual algorithms are all built on the same scheme: (i) extraction and combination of features (low 
level features from millions of pixels), and (ii) intelligent algorithms on higher level of features. 
These algorithms are not directly related to the machine architecture.
You can implement in your favorite programming language these algorithms on a mono-processor, on a multi-core processor or even on a cluster of machines. 
The critical point of all these algorithms is having to {\em explicitly} create from raw data feature elements.

Although the idea of using neural network architectures is not new in itself, a great revolution took place in visual processing through deep learning techniques. 
Deep learning~\cite{DL-2015} (DL) proceeds by {\em stacking} layer by layer basic artificial neural networks (technically called restricted Boltzmann machines, RBMs). 
A basic neural network consists of only two layers: A layer of {\em visible neurons} connecting the neurons to the input values,  
 and a layer of {\em hidden neurons} (not seeing directly the input) but evaluation a function combining the values of ​​visible neurons (there is no connection between the hidden neurons). 
Learning restricted Boltzmann machines (shallow network) is easy, and we can then build larger neural machinery by stacking these limited two-layer machines:
The hidden layer becomes  the visible layer of the upper neural network. 
There are several basic types of neuron network following the model of the chosen neuron (long short-term memory recurrent neural network, etc.) and type of connections between neurons.

Thus in  deep learning, features are {\em  implicitly} learned by the neural network itself, directly  from the input data, and do not need to be engineered. 
Information is extracted by stacked layers of features: At the low level, we observe features close to characteristics  observed in the brain (like Gabor filters) and at higher levels, features that correspond to concepts more abstract that have an increasing semantic meaning for the visual task. 
Today we can learn neuron networks with twenty layers or so. 
This learning technique is called ``Deep Learning''   to contrast with shallow learning that requires to engineer by hand the features of an image.

\section{Summary}
Digital imaging is an exciting area that goes far beyond the traditional manipulation photo pictures: 
Researchers are always seeking for novel imaging modalities that will shed light on key issues in Science (for example, in astronomy or in cell biology).
The information stored in an image is not a well-defined concept but rather a concept that depends on the visual task to solve at hand:
 For simple tasks (like detecting a red ball by a soccer robot), the image information to be extracted is concise, 
but for more complex tasks such as understanding road traffic conditions  from a camera mounted on a car, the information to be extracted is much larger, need to be fused, and further depends on  auxiliary information provided by the problem domain knowledge (for example, when detecting road lanes use the known dimensions of lane patterns, a prior knowledge, to improve the recognition rate).

\section*{Acknowledgments}
The author is very grateful for the kind invitation of Professor Gilles Dowek and Professor Dominique Rojat, and acknowledges the support of Ms. B\'eatrice Salviat of La main \`a la p\^ate.

{\small
\section*{Figure credits}
 \begin{tabular}{ll}
Figure~\ref{fig:bayer} & \url{https://en.wikipedia.org/wiki/Demosaicing} \\
 & \protect\url{https://en.wikipedia.org/wiki/Ultrasound} \\
Figure~\ref{fig:ultrasound} &\url{https://en.wikipedia.org/wiki/Magnetic_resonance_imaging}\\
Figure~\ref{fig:captcha}& \url{https://en.wikipedia.org/wiki/ReCAPTCHA} \\
Figure~\ref{fig:robotcup}&\url{Aibos_playing_football_at_Robocup_2005.jpg}\\
Figure~\ref{fig:smileshutter}& \url{https://www.sony.com.au/product/resources/en_AU/}\\
& \url{images/Technology/DI/Camera/illustration/smile_shutter.jpg}
\end{tabular}{ll}
}


\end{document}